\pdfoutput=1

\documentclass[11pt]{article}

\usepackage[final]{acl}

\usepackage{times}
\usepackage{latexsym}
\usepackage{bm}
\usepackage{appendix}
\usepackage{multirow}
\usepackage{booktabs}
\usepackage{amssymb}
\usepackage{tabularx}

\usepackage[T1]{fontenc}

\usepackage[utf8]{inputenc}

\usepackage{microtype}

\usepackage{inconsolata}
\usepackage{amsmath}
\usepackage{graphicx}
\usepackage{cleveref}
\title{See More, Store Less: Memory-Efficient Resolution for Video Moment Retrieval}

\author{
    Mingyu Jeon\textsuperscript{\rm 1},
    Sungjin Han\textsuperscript{\rm 1},
    Jinkwon Hwang\textsuperscript{\rm 1},
    Minchol Kwon\textsuperscript{\rm 1},
    Jonghee Kim\textsuperscript{\rm 2},
    Junyeong Kim\textsuperscript{\rm 1}\\
    \textsuperscript{\rm 1}Department of Artificial Intelligence, Chung-Ang University\\
    \textsuperscript{\rm 2}Electronics and Telecommunications Research Institute (ETRI)\\
    \small{\{smart2557, sungjinhan, wlsrnjs905, welchs3576, junyeongkim\}@cau.ac.kr, jhkim27@etri.re.kr}}

\begin{document}
\maketitle
\begin{abstract}
Recent advances in Multimodal Large Language Models (MLLMs) have improved image recognition and reasoning, but video-related tasks remain challenging due to memory constraints from dense frame processing. Existing Video Moment Retrieval (VMR) methodologies rely on sparse frame sampling, risking potential information loss, especially in lengthy videos. We propose SMORE (\textbf{S}ee \textbf{MORE}, store less), a framework that enhances memory efficiency while maintaining high information resolution. SMORE  (1) uses query-guided captions to encode semantics aligned with user intent, (2) applies query-aware importance modulation to highlight relevant segments, and (3) adaptively compresses frames to preserve key content while reducing redundancy. This enables efficient video understanding without exceeding memory budgets. Experimental validation reveals that SMORE achieves state-of-the-art performance on QVHighlights, Charades-STA, and ActivityNet-Captions benchmarks. 
\end{abstract}


\section{Introduction}
\label{sec:intro}

\begin{figure}[t]
\centerline{\includegraphics[width=1\columnwidth]{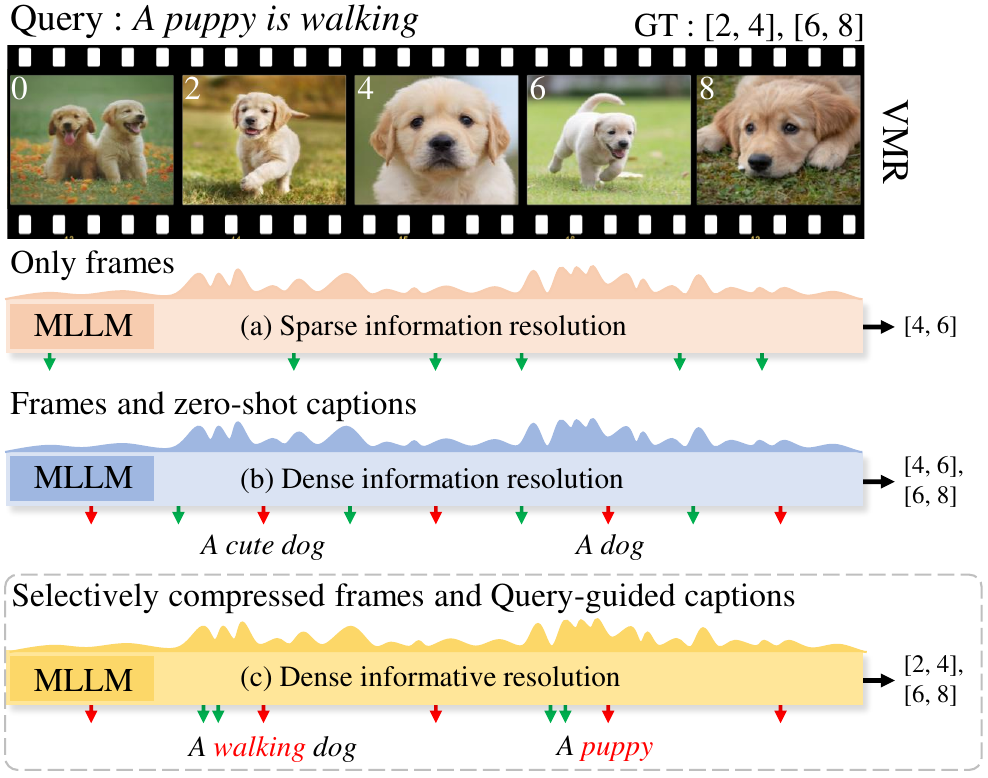}}
\caption{ Illustrates how MLLMs can be applied to VMR while considering a notion of \emph{information resolution}:
\textbf{(a)} Random or sparse sampling may miss key scenes due to low information resolution.
\textbf{(b)} Zero-shot captioning offers dense coverage but lacks user intent, leading to less relevant captions for retrieval.
\textbf{(c)} Our proposed SMORE yields dense and informative representation, preserving key details with query awareness.}
\label{fig_1}
\end{figure}

Multimodal Large Language Models (MLLMs) have significantly advanced video understanding across a range of tasks, including video question answering, video summarization, and video moment retrieval.
The task of Video Moment Retrieval (VMR) focuses on identifying and retrieving temporal segments within video content that semantically correspond to a given linguistic query. VMR is particularly challenging among video-language tasks due to its need for fine-grained temporal reasoning and high-precision alignment between visual content and natural language queries.

Recent VMR systems are frequently constrained by limited memory resources, as they must process densely sampled video frames over extended temporal durations. Notably, such memory bottlenecks are not exclusive to VMR but also arise in other video-language tasks, including Video Question Answering (VQA). To address these limitations, recent studies in VQA have proposed two primary strategies: (1) leveraging textual captions to abstract visual content, and (2) reducing temporal redundancy within frame sequences.
First, Caption-based approaches attempt to reduce memory load by condensing visual content into short textual representations. 
However, conventional captioning frameworks typically generate generic descriptions, often resulting in inherently sub-optimal captions due to their lack of conditioning on user intent.
Second, existing redundancy reduction methods primarily rely on keyframe sampling, which often compromises temporal fidelity and makes them unsuitable for VMR tasks that require precise temporal alignment.

While the above strategies have proven effective in VQA tasks, they cannot be directly applied to VMR due to fundamental differences. In VQA, generic descriptions or temporally sparse representations often suffice to identify relevant information. In contrast, VMR requires fine-grained alignment between visual content and user intent to accurately localize events within the video timeline.

To address these limitations, we propose \textbf{SMORE} (See MORE, store less), a novel framework that enhances query-video alignment and reduces visual redundancy. This is achieved through two key mechanisms: (1) memory-efficient query-video alignment via query-guided caption generation, where semantic relevance is further refined through query-aware importance scoring; and (2) structured visual compression that produces a compact set of informative visual embeddings, preserving rich visual evidence with reduced redundancy.

The proposed query-guided captioning module enables the linguistic query to directly inform caption generation. 
This paradigm refines semantic representations to better align with the query, thereby improving retrieval precision. 
Specifically, our method first filters scenes through query-relevance classification using question answering (QA)-based prompts, and then generates captions with query-guided prompts to ensure semantic alignment. 
To further enhance alignment, we assign an importance score to each frame-caption pair based on query-video-caption similarity, allowing the LLM to focus on the most relevant segments.

To improve memory efficiency, we propose a structured visual compression strategy motivated by the observation that not all video frames contribute equally to semantic understanding. By measuring inter-frame visual similarity, we identify and compress redundant frames while preserving diverse frames at high resolution, enabling compact yet informative visual representation.

Our proposed SMORE exhibits strong performance even when operating within a restricted memory environment using an A6000 GPU with 48GB of memory. 
In comparison, recent methodologies, Chrono and LLaVA-MR, have utilized an A100 GPU with 80GB of memory. 
Despite such constraints, our method achieves a 3.35\% mAP average improvement over Chrono on the QVHighlights benchmark and outperforms the current state-of-the-art model, SG-DETR, by 4.19\% on R1@0.5.
Moreover, our framework consistently achieves superior performance across all the evaluation metrics on the Charades-STA and ActivityNet-Captions benchmarks.

\section{Related work}
\subsection{Video Moment Retrieval}


\begin{figure*}[t]
\centerline{\includegraphics[width=1\textwidth]{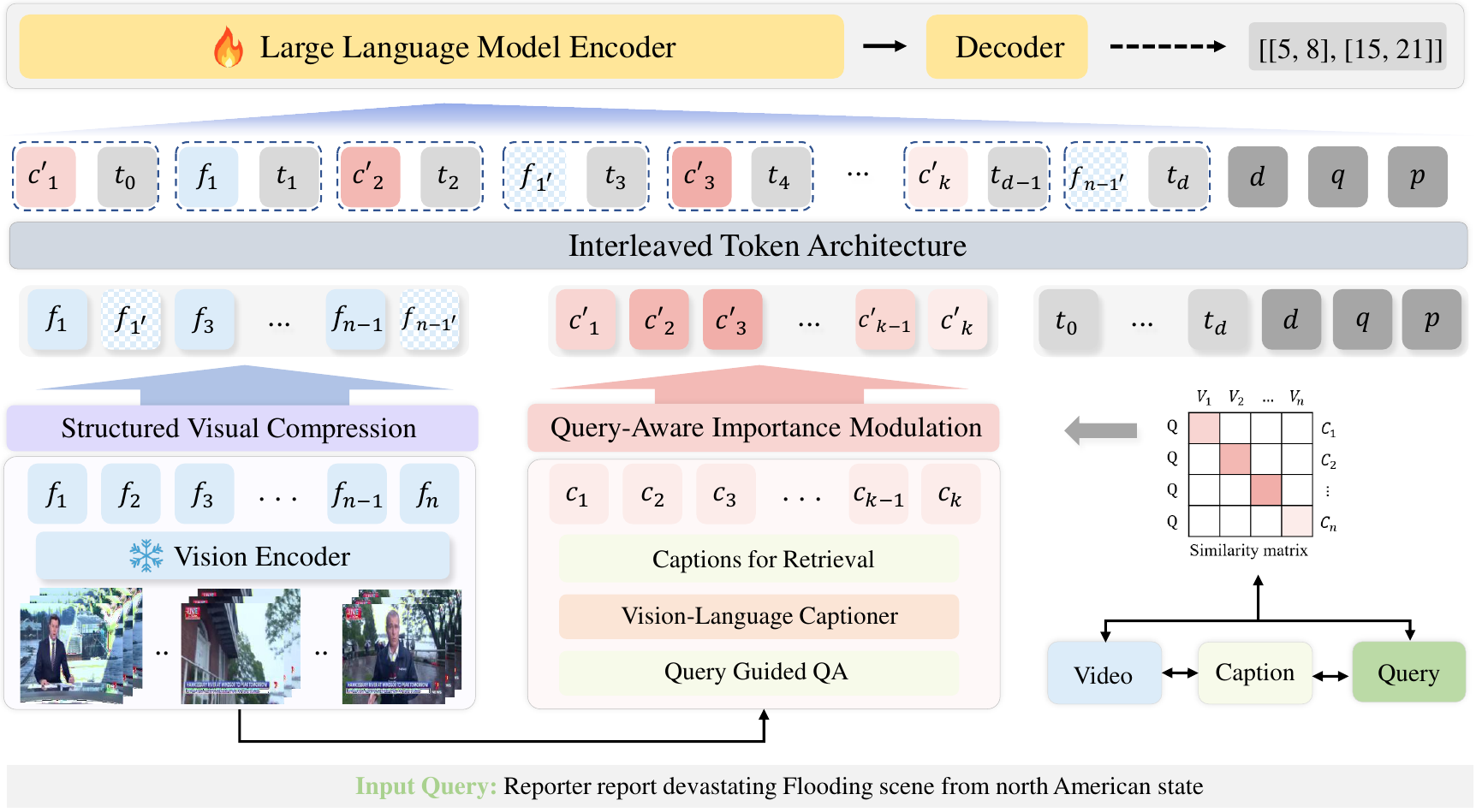}}
\caption{The overall architecture of SMORE.
It first generates query-guided captions through QA (Sec.\ref{subsec: Query-guided Captions for Retrieval}). 
Next, query-aware importance modulation adjusts the relative importance between frames, captions, and queries (Sec.\ref{subsec: Query-Aware Importance Modulation}).
By considering the information resolution, we efficiently reduce redundant information among the frame embeddings from the vision encoder (Sec.\ref{subsec: Structured Frame Compression}). 
The LLM encoder maps these frame tokens $f$ and caption tokens $c$ to their corresponding time embedding $t$ and interleaves them as input. Finally, the decoder outputs the temporal segment corresponding to the query.}
\label{fig:method}
\end{figure*}


Video Moment Retrieval (VMR) aims to accurately extract relevant temporal segments from a video based on a natural language query.
Early VMR methods~\cite{Charades,2017localizing} relied on fixed candidate generation techniques, such as sliding windows and temporal anchors, but suffered from computational inefficiency due to complex pre- and post-processing.
Transformer-based models, such as Moment-DETR~\cite{QVH}, introduced end-to-end set prediction, which improved retrieval performance~\cite{35_QD-DETR,bam-detr}. However, these models often require extensive pretraining and fixed prediction structures.

Recent VMR studies using MLLMs have introduced approaches such as SeViLA~\cite{33_SeViLA}, Chrono~\cite{13_chrono}, and LLaVA-MR~\cite{14_LLaVA_MR}. 
SeViLA employs sparse keyframe selection for moment retrieval but struggles to fully capture a video's temporal context. 
Its performance heavily relies on selecting the right keyframes, which can result in incomplete representations when crucial moments are missed. 
Chrono and LLaVA-MR enhance temporal awareness and retrieval accuracy, yet the increase in visual information significantly raises memory consumption.
Our proposed SMORE framework improves memory efficiency and retrieval precision by incorporating query-guided captioning and query-aware importance modulation.


\subsection{Information Compression for MLLMs}

Memory efficiency is a crucial research topic when applying Vision-Language Models (VLMs).
Researchers actively explore methods to compress redundant information and eliminate unnecessary frame data when processing lengthy videos~\cite{18_MovieChat, LongVU,9_LLoVI,20_ChatUniVi,19_SimpleFrames}.
In previous studies, researchers attempted to reduce video frame-level redundancy by replacing frames with captions for summarization or merging consecutive redundant frames.
Similar approaches have also been extensively studied in image processing, where only key information is retained while less relevant details are compressed~\cite{visionzip, SparseVLM}. 
Although these methods significantly reduce memory usage by eliminating unnecessary information, they often struggle to preserve temporal order and segment information within videos.
Building on these compression strategies, we propose a method that preserves key information while effectively compressing redundant content.

\section{Methods}

\subsection{Overview}
In conventional MLLM-based VMR~\cite{13_chrono}, the input structure consists of video frames ($f_i$) interleaved with their corresponding timestamps ($t_j$), relying on sparse sampling due to memory constraints. 
We propose SMORE, which incorporates query-guided captions into video frames within the interleaved structure. As shown in Figure~\ref{fig:method}, this format supports richer, intent-aligned representations by enhancing semantic coverage using lightweight textual captions instead of increasing the number of visual tokens, thereby maintaining memory efficiency.
This enriched sequence is then augmented with video duration metadata, the retrieval query, and an instructional prompt to guide the MLLM's reasoning.

Building on this formulation, SMORE incorporates training-free components to achieve semantically expressive and memory-efficient modeling: 

\noindent (1) \textbf{Query-guided caption generation} (Section~\ref{subsec: Query-guided Captions for Retrieval}), which generates a set of captions \( \mathcal{C}=\{c_1, \cdots, c_k\} \), where \( c_i \in \mathcal{C} \) is aligned with the retrieval query. 

\noindent (2) \textbf{Query-aware importance modulation} (Section~\ref{subsec: Query-Aware Importance Modulation}), which adjusts the relevance of the captions by transforming them into weighted representations \( \mathcal{C}'=\{c_1', \cdots, c_k'\} \), where \( c_i' \in \mathcal{C}' \) denotes the re-weighted embedding of \( c_i \in \mathcal{C} \).
(3) \textbf{Structured visual compression} (Section~\ref{subsec: Structured Frame Compression}), 
which defining the compressed frame sequence \( \mathcal{F}' \) by conditionally replacing redundant frame embeddings in \( \mathcal{F} = \{f_1, \cdots, f_n\} \) with their compressed frame representations \( f_i' \).


\subsection{Query-Guided Caption Generation}
\label{subsec: Query-guided Captions for Retrieval}


\begin{figure}[h]
\centerline{\includegraphics[width=1\columnwidth]{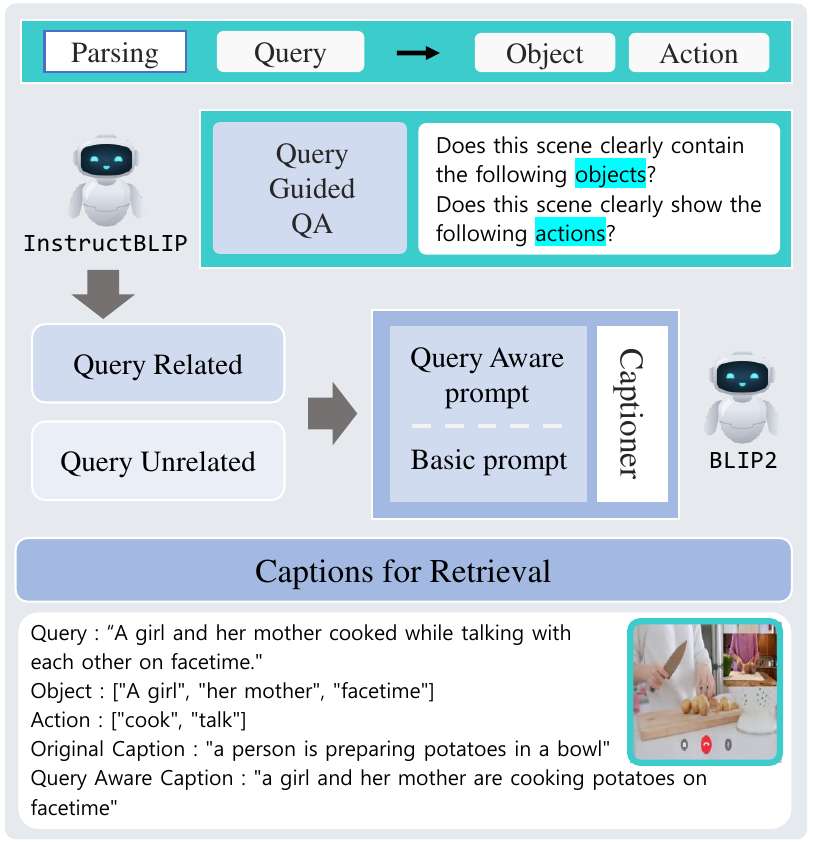}}
\caption{Query-guided caption generation. The query is parsed into objects and actions, which guide a QA-based relevance check. Relevant scenes receive query-aware prompts for captioning, improving alignment with retrieval goals.}
\label{fig:query-guided caption}
\end{figure}


Current VMR methods~\cite{13_chrono,14_LLaVA_MR} demand substantial GPU memory proportional to the number of selected frames and risk missing critical information due to irregular sampling. For example, extracting 60 frames from a 150-second video may yield gaps of up to 5 seconds between frames, making it difficult to capture ephemeral visual cues.
To mitigate such information loss and construct denser video representations, we incorporate \textit{zero-shot captioning}, which generates textual descriptions for densely divided temporal segments. 
These captions can be efficiently processed by LLMs and help enhance semantic coverage without increasing visual token count.

While zero-shot captioning facilitates comprehensive video understanding, it exhibits inherent limitations regarding contextual alignment with user queries.
Specifically, even meticulously detailed captions may lack relevance to the user's retrieval objectives if the captioning process fails to prioritize query-specific information.

To overcome this challenge, we introduce Query-Guided Captioning, a novel approach that aligns the video description process with user-specific retrieval goals. Our method substantially enhances retrieval performance through two primary advantages: 1) it generates more semantically meaningful and contextually relevant captions for query-aligned scenes, and 2) it prevents the generation of distracting or misleading captions for irrelevant scenes. This is achieved by first analyzing the objects and actions present in the user query to understand its core intent.

Based on the query, we perform a simple question-answering (QA)-based classification to evaluate the relevance of each scene by answering: \texttt{"Does this object/action appear in the scene?"}. Only scenes that satisfy these criteria proceed to the query-guided captioning phase, where we employ a query-aware prompt (e.g., \texttt{"Generate a caption that is relevant to the query"}). Conversely, scenes that fail the check proceed to the original caption generation phase. This dual-path strategy ensures that detailed, query-focused descriptions are generated only when necessary, maximizing the overall relevance of the captions for effective retrieval.


\subsection{Query-Aware Importance Modulation}
\label{subsec: Query-Aware Importance Modulation}

In video retrieval and understanding tasks, not all captions contribute equally; some convey essential content, while others include background or redundant information. Treating all captions uniformly may dilute key information relevant to the query. To address this, we introduce a weighting mechanism based on the semantic similarity between the query, video frames, and generated captions, enabling the model to focus on the most informative content and improve retrieval performance.
Specifically, we formulate a similarity-based caption weighting score \( S_{q}(f_i, c_i) \) as follows:
\begin{equation}
    S_{q}(f_i, c_i) = \alpha_1 \, V(f_i, q) + \alpha_2 \, \overline{V}(q, f_i, c_i)
    \label{eq:caption_weight}
\end{equation}
Here, \( q \in \mathcal{Q} \) denotes the embedding of the textual query,  
\( f_i \in \mathcal{F} \) represents the embedding of the \( i \)-th video frame,  
and \( c_i \in \mathcal{C} \) is the embedding of the corresponding caption.
The term \( V(f_i, q) \) measures the visual-query similarity via cosine similarity between the frame and the query embeddings.  
The term \( \overline{V}(q, f_i, c_i) \) refines the query-caption similarity \( V(q, c_i) \) by incorporating the frame-caption similarity \( V(f_i, c_i) \), inspired by the CLIPScore~\cite{CLIPScore} formulation.
The coefficients \( \alpha_1 \) and \( \alpha_2 \) are hyperparameters that balance the contribution of each component.
Before entering the LLM's self-attention layer, each caption embedding \( c_i \) is re-weighted using its relevance score \( S_q(f_i, c_i) \):
\begin{equation}
c_i' = S_q(f_i, c_i) \cdot c_i
\end{equation}
%
%
\begin{figure}[h]
\centerline{\includegraphics[width=1\columnwidth]{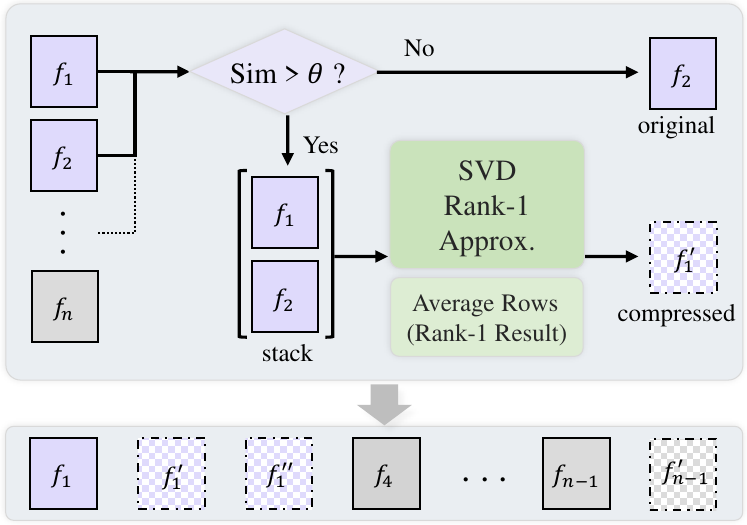}}
\caption{Illustration of the structured visual compression module. 
Redundant frame pairs are identified via cosine similarity and compressed using truncated SVD to produce compact, information-preserving embeddings.}
\label{fig_4}
\end{figure}
%
This yields a set of re-weighted embeddings \( \mathcal{C}' \), where each \( c_i' \in \mathcal{C}' \) is modulated by its semantic relevance, conceptually similar to the softmax-based weighting in Transformer attention~\cite{transformer}, to prioritize more informative captions. Furthermore, this modulation helps mitigate issues arising from ambiguous queries; by down-weighting irrelevant information, it reduces the risk of incorrect outputs based on spurious correlations.


\subsection{Structured Visual Compression}
\label{subsec: Structured Frame Compression}

Video frequently contains a sequence of highly similar frames that introduce memory inefficiency through information redundancy. In the context of VMR, efficiently handling such redundancy is essential, allowing models to concentrate on significant temporal and contextual features.

To address frame-level redundancy, we propose a simple yet effective approach termed \textit{Structured Visual Compression (SVC)}. 
This low-overhead mechanism more effectively maintains temporal information than simple redundancy reduction methods like frame sampling, while reducing spatial redundancy more effectively compared to average pooling. By employing a straightforward yet powerful SVD-based approach, it preserves essential spatial characteristics by retaining dominant components, a process that effectively discarding lower-order redundancies and allowing the model to focus on semantically rich content.

Specifically, frame embeddings are processed sequentially, with the first frame in the video acting as the initial anchor \( f_a \in \mathbb{R}^D \), where $D$ is the dimension of the frame embeddings. For each subsequent frame \( f_i \in \mathbb{R}^D \), we compute the cosine similarity with the current anchor. If the similarity exceeds a predefined threshold \( \theta \), indicating redundancy, the two frames are stacked into a temporary composite representation \( M_i \), which is then compressed using truncated singular value decomposition (SVD). Otherwise, \( f_i \) is retained as is and becomes the new anchor for subsequent comparisons.

Specifically, the selection rule is defined as:
\begin{equation}
f_i =
\begin{cases}
M_i, & \text{if } \text{sim}(f_a, f_i) > \theta \\
f_i, & \text{otherwise}
\end{cases}
\label{eq:similarity_condition}
\end{equation}

Here, \( M_i \) denotes the stacked embedding of the anchor and current frames:
\begin{equation}
M_i = \begin{bmatrix} f_a \\ f_i \end{bmatrix}
\label{eq:stack_matrix}
\end{equation}


\begin{table*}[t]
\centering
\scriptsize
\renewcommand{\arraystretch}{1.1}
\resizebox{\textwidth}{!}{%
\begin{tabular}{l l c c c  c c  c c c c c}
\toprule
\multirow{2}{*}{\textbf{Method}} & \multirow{2}{*}{\textbf{Venue}}
  & \multicolumn{5}{c}{\textbf{Test}} 
  & \multicolumn{5}{c}{\textbf{Validation}} \\
\cmidrule(lr){3-7} \cmidrule(lr){8-12}
 &  & \textbf{R1@0.5} & \textbf{R1@0.7} & \textbf{mAP@0.5} & \textbf{mAP@0.75} & \textbf{Avg.}
      & \textbf{R1@0.5} & \textbf{R1@0.7} & \textbf{mAP@0.5} & \textbf{mAP@0.75} & \textbf{Avg.} \\
\midrule
Moment-DETR \cite{QVH} & NeurIPS & --     & --     & --     & --     & --   
                       & 59.68  & 40.84  & --     & --        & 32.2   \\
EaTR \cite{37_EaTR}    & ICCV    & --     & --     & --     & --     & --   
                       & 61.36  & 45.79  & 61.86  & 41.91     & \underline{41.74} \\
QD-DETR \cite{35_QD-DETR}  & CVPR    & 62.40  & 44.98  & 62.52  & 39.88  & 39.86
                       & 62.68  & 46.66  & 62.23  & 41.82     & 41.22  \\
UnLoc-L \cite{38_UnLoc-L}  & ICCV    & --     & --     & --     & --     & --   
                       & 66.10  & 46.70  & --     & --        & --     \\
Chrono \cite{13_chrono}    & --   & 74.77  & 60.51  & 68.12  & 53.38  & 51.37
                       & 76.13  & 63.35  & 69.39  & 55.78 & --     \\
LLaVA-MR \cite{14_LLaVA_MR} & --   & \textbf{76.59} & \underline{61.48} 
                            & 69.41  & 54.40  & 52.73
                            & \underline{78.13} & \underline{64.13} 
                            & \underline{69.64} & \textbf{56.32} & --     \\
SeViLa \cite{33_SeViLA}     & NeurIPS & 54.50  & 36.50  & --     & --     & 32.30
                            & --     & --     & --     & --        & --     \\
UniVTG \cite{34_UniVTG}     & ICCV    & 58.86  & 40.86  & 57.60  & 35.59  & 35.47
                            & --     & --     & --     & --        & --     \\
InterVideo2-6B \cite{40_InternVideo2}  & ECCV    & 71.42  & 56.45  & --     & --     & 49.24
                            & --     & --     & --     & --        & --     \\
SG-DETR \cite{41_SG-DETR}   & --   & 72.20  & 56.60  & \textbf{73.20} & \underline{55.80} & \underline{54.10}
                            & --     & --     & --     & --        & --     \\
\midrule
\textbf{SMORE (Ours)}       & --      & \underline{76.39} & \textbf{62.84} 
                            & \underline{70.40} & \textbf{55.90} & \textbf{54.72}
                             & \textbf{78.84}     & \textbf{64.19}      
                            & \textbf{70.13}     & \underline{55.89}            & \textbf{54.25} \\
\bottomrule
\end{tabular}}
\caption{QVHighlights Performance comparison of various methods based on multiple metrics including MR-full-R1, MR-full-mAP. The ``-" in the Venue column indicates that the work is unpublished.}
\label{tab:1}
\end{table*}


To compress redundant pairs, we apply rank-\( k \) truncated SVD:
\begin{equation}
M_i \approx  U_k \Sigma_k V_k^\top
\label{eq:rankk_svd}
\end{equation}

We then average the result to get compressed frame representation \( f_i' \in \mathbb{R}^D \), a compact embedding that captures the dominant semantics of the pair, contributing to the compressed frame sequence \( \mathcal{F}' \).


\begin{table}[t!]
\renewcommand{\arraystretch}{1.05}
\centering
\scriptsize
\resizebox{1\columnwidth}{!}{%
\begin{tabular}{l l c c c}
\toprule
\multirow{2}{*}{\textbf{Method}} & \multirow{2}{*}{\textbf{Venue}} 
    & \multicolumn{3}{c}{\textbf{R1}} \\
\cmidrule(lr){3-5}
 &  & \textbf{mIoU} & \textbf{@0.5} & \textbf{@0.7} \\
\midrule
Moment-DETR~\cite{QVH}        & NeurIPS        & -     & 53.63 & 31.37 \\
QD-DETR~\cite{35_QD-DETR}     & CVPR           & -     & 57.31 & 32.55 \\
UniVTG~\cite{34_UniVTG}       & ICCV           & 50.1  & 58.01 & 35.65 \\
UnLoc-L~\cite{38_UnLoc-L}     & ICCV        & -     & 60.80 & 38.40 \\
UniMD+Sync~\cite{39_UniMD}    & ECCV          & -     & 63.98 & 44.46 \\
InternVideo2-1B~\cite{40_InternVideo2} & ECCV   & -     & 68.36 & 45.03 \\
EaTR~\cite{37_EaTR}           & ICCV          & -     & 68.47 & 44.92 \\
Chrono~\cite{13_chrono}       & --        & 58.63 & 69.31 & 49.29 \\
InternVideo2-6B~\cite{40_InternVideo2} & ECCV   & -     & 70.03 & 48.95 \\
SG-DETR~\cite{41_SG-DETR}     & --        & 59.1  & 70.20 & 49.50 \\
LLaVA-MR~\cite{14_LLaVA_MR}   & --           & \underline{59.78} & \underline{70.65} & \underline{49.58} \\
\midrule
\textbf{SMORE(Ours)}                & --               & \textbf{60.9} & \textbf{71.26} & \textbf{49.81} \\
\bottomrule
\end{tabular}}
\caption{Performance comparison on Charades-STA based on mIoU, R1@0.5 and R1@0.7 metrics.}
\label{tab:2}
\end{table}


\section{Experiments}

We validate our proposed SMORE on three representative video moment retrieval (VMR) datasets: QVHighlights~\cite{QVH}, Charades-STA~\cite{Charades}, and ActivityNet-Captions~\cite{ActivityNet}. 

\noindent\textbf{Datasets}
\textbf{(1)} \textbf{QVHighlights} is a large-scale dataset for text-based moment retrieval, containing over 10,000 YouTube videos with long, complex queries. Performance is evaluated on a hidden test set via an official server.
\textbf{(2)} \textbf{Charades-STA} features nearly 10,000 videos with 16,128 annotations, focusing on moment localization from sentences in shorter, everyday activity videos.
\textbf{(3)} \textbf{ActivityNet-Captions} is a large-scale benchmark consisting of 20,000 untrimmed videos with approximately 100,000 captions, widely used for both moment retrieval and dense-event captioning.
%


\noindent\textbf{Evaluation Metrics.}
Model performance is evaluated using standard VMR metrics such as Recall@K and mean Average Precision (mAP).
The metric Recall@K measures the proportion of queries for which at least one of the top-K predicted segments exceeds a certain IoU threshold. 
For example, R1@0.5 refers to the Recall@1 performance when the prediction includes a ground truth segment with an IoU of at least 0.5. 
The metric mAP measures the average precision at a specified IoU threshold, typically reported as mAP@0.5 and mAP@0.75.


\begin{table}[t]
    \centering
    \renewcommand{\arraystretch}{1}
    \resizebox{0.9\columnwidth}{!}{%
    \begin{tabular}{l l c c}
        \toprule
        \textbf{Method} & \textbf{Venue} & \textbf{R1@0.5} & \textbf{R1@0.7} \\
        \midrule
        DRN \cite{DRN}             & CVPR   & 45.45 & 24.36 \\
        UnLoc-L \cite{38_UnLoc-L}  & ICCV   & 48.30 & 30.20 \\
        Chrono \cite{13_chrono}    & --  & 53.92 & 35.55 \\
        LLaVA-MR \cite{14_LLaVA_MR}& --  & \underline{55.16} & \underline{35.68} \\
        NumPro-FT \cite{NumPro}    & CVPR   & 37.50 & 20.60 \\
        \midrule
        \textbf{SMORE (Ours)}      & --     & \textbf{56.31}     & \textbf{36.34} \\
        \bottomrule
    \end{tabular}}
    \caption{Performance comparison on ActivityNet-Captions based on R1@0.5 and R1@0.7.}
    \label{tab:3}
\end{table}


\subsection{Implementation Details}
Feature similarity was computed using a CLIP-based model~\cite{CLIP}. For language modeling, we used Flan-T5 XL~\cite{7_Flant5}, fine-tuned via LoRA~\cite{LoRA} on 0.6266\% of parameters. InstructBLIP~\cite{8_InstructBLIP} was used for query-guided captioning, and BLIP2~\cite{blip2} for general captioning.
Further details are provided in the Appendix.


\subsection{Quantitative Results}

A fundamental challenge in VMR evaluation is the inherent trade-off between R@1 and mAP. R@1 can be inflated by long predictions under loose IoU thresholds, while mAP demands precise localization, penalizing segmentation errors.


\begin{table}[h]
\centering
\renewcommand{\arraystretch}{1.1}
\resizebox{1\columnwidth}{!}{ 
\begin{tabular}{lcccc}
\toprule
\textbf{Component} & \textbf{mIoU} & \textbf{R1@0.5} & \textbf{mAP@0.5} & \textbf{mAP Avg.} \\
\midrule
(a) Baseline Only                          & 69.93 & 75.87 & 68.27 & 51.39 \\
(b) + Zero-shot Captioning                 & 70.68 & 76.63 & 68.15 & 52.64 \\
(c) + Query-guided Captioning         & 71.60 & 76.02 & 69.97 & 53.38 \\
(d) + Importance Modulation               & 72.35 & 77.74 & 69.15 & 53.78 \\
(e) + Structured Visual Compression         & \textbf{72.61} & \textbf{78.84} & \textbf{70.13} & \textbf{54.25} \\
\bottomrule
\end{tabular}}
\caption{Ablation study showing the effect of each component.}
\label{tab:4}
\end{table}


\begin{table}[h]
\centering
\renewcommand{\arraystretch}{1.1}
\resizebox{1\columnwidth}{!}{ 
\begin{tabular}{lccccc}
\toprule
\textbf{Compression Method} & \textbf{R1@0.5} & \textbf{R1@0.7} & \textbf{mAP@0.5} & \textbf{mAP@0.75} & \textbf{mAP Avg.} \\
\midrule
Frame Selection & 73.94 & 59.81 & 66.3  & 52.46 & 51.02 \\
Average Pooling & 75.42 & 61.42 & 68.1  & 53.96 & 52.77 \\
\textbf{SVD (Ours)} & \textbf{78.84} & \textbf{64.19} & \textbf{70.13} & \textbf{55.89} & \textbf{54.25} \\
\bottomrule
\end{tabular}}
\caption{Ablation study on the method for Structured Visual Compression (SVC).}
\label{tab:5}
\end{table}


Consequently, prior methods typically excel at either R@1 or mAP, but not both. In contrast, our method surpasses all baselines on both metrics, demonstrating a superior balance between retrieval accuracy and localization precision.

As presented in Table~\ref{tab:1}, SMORE demonstrates its superiority on the official QVHighlights test set by addressing the common trade-off between mAP and recall that affects leading models. Specifically, SMORE improves upon the high-mAP model, SG-DETR~\cite{41_SG-DETR}, with significant recall gains of +4.19\% (R1@0.5) and +6.24\% (R1@0.7), while also maintaining a higher mAP average (+0.62\%). Furthermore, it surpasses the high-recall model, LLaVA-MR~\cite{14_LLaVA_MR}, with a 1.36\% improvement in R1@0.7 and a 1.99\% higher mAP average.

We further demonstrate SMORE's effectiveness on the Charades-STA and ActivityNet-Captions datasets, where it consistently outperforms both baseline and state-of-the-art (SOTA) models.
%
\begin{figure}[h]
\centerline{\includegraphics[width=1\columnwidth]{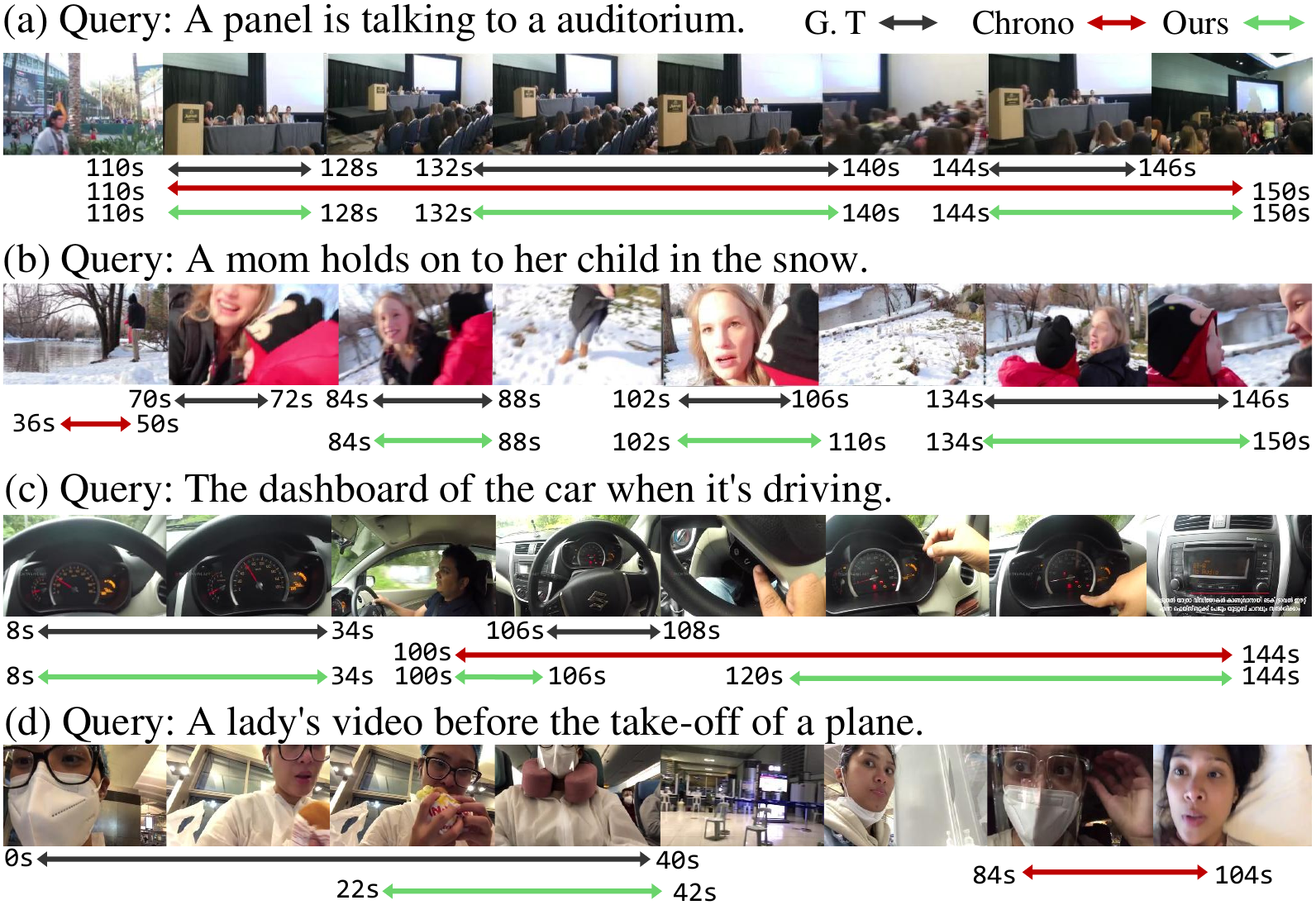}}
\caption{Qualitative results on the QVHighlights datasets.}
\label{fig_5}
\end{figure}
As shown in Table~\ref{tab:2}, on the Charades-STA dataset, SMORE achieves a new SOTA. It surpasses LLaVA-MR with a 1.12\% improvement in mIoU and 0.61\% in R1@0.5. The performance gain is even more significant compared to the Chrono baseline, with mIoU increasing by 2.27\%.
Similarly, on the ActivityNet-Captions dataset (Table~\ref{tab:3}), SMORE again sets a new performance benchmark. It outperforms the previous SOTA model, LLaVA-MR, with gains of 1.15\% in R1@0.5 and 0.66\% in R1@0.7.

SMORE achieves a new state-of-the-art performance across both short and long video benchmarks, including QVHighlights, Charades-STA, and ActivityNet-Captions, while using less memory than standard MLLM models (as analyzed in Section~\ref{Memory}).

\subsection{Ablations}

To precisely quantify the contribution of each component, we conducted a progressive ablation study on QVHighlights under a 48GB memory constraint, with the results detailed in Table~\ref{tab:4}. We consistently tracked key metrics like mIoU and mAP Average to evaluate the marginal gain at each stage.

Our analysis begins with the (a) baseline, which uses only video frames and achieves an mAP Average of 51.39\%. First, by incorporating (b) zero-shot captions, the mAP Average improves by +1.25\% to 52.64\%, demonstrating that even query-agnostic semantic context is highly beneficial.
Next, replacing these with (c) query-guided captions yields an additional +0.74\% gain in mAP Average, confirming that aligning captions with the query is crucial for performance. Building on this, (d) query-aware importance modulation provides a further +0.40\% mAP Average improvement, which validates the effectiveness of guiding the model's focus toward relevant information. Finally, applying (e) structured visual compression provides the last performance lift, adding another +0.47\% to the mAP Average by reducing redundancy and emphasizing key visual moments.
Cumulatively, the full SMORE model (e) achieves a total improvement of +2.68\% in mIoU, +2.97\% in R1@0.5, and +2.86\% in mAP Average over the initial baseline (a). This step-by-step analysis validates that each proposed component provides a distinct and synergistic contribution to the final performance.

Furthermore, we conducted an additional ablation study to validate our specific design choice for the Structured Visual Compression (SVC) module. As shown in Table~\ref{tab:5}, we compared our SVD-based approach against two common alternatives: naive frame selection, a method that completely discards redundant frames along with their temporal information, and average pooling. The results clearly indicate that our SVD-based method is the most effective, outperforming the next-best approach, average pooling, by +1.48\% in mAP Average.

This performance gain validates our hypothesis. Unlike simple frame selection, which risks discarding critical temporal information, our SVD-based approach effectively captures rich visual dynamics. Moreover, it preserves essential spatial characteristics by retaining dominant components, a capability that is often diluted by simple average pooling. This allows the model to focus on semantically rich content while efficiently reducing redundancy.

\subsection{Qualitative Results}

Figure~\ref{fig_5} presents four qualitative examples evaluating the predictive performance of SMORE. 
First, (a) shows that our model effectively predicts the boundaries of various segments, yielding results that closely align with the ground truth. 
In (b), while the baseline model predicted the segment where a man is holding a child as the correct interval, SMORE enhances prediction precision by leveraging subtle cues from the QA module during caption generation.
In contrast, (c) shows a case where the predicted segment is broader than the ground truth. This can be understood as a result of an ambiguous situation. The ambiguity is caused by the cameraman's hand shaking so much that it is mistaken for the movement of a car in the query. 
\begin{table}[h!]
\centering
\renewcommand{\arraystretch}{1} %
\resizebox{1\columnwidth}{!}{%
\begin{tabular}{lccc|cccccc}
\toprule
{\textbf{MEM}} & \multicolumn{3}{c}{\textbf{Baseline}} & \multicolumn{3}{c}{\textbf{SMORE}} \\
\cmidrule{2-7}
& \textbf{R1@0.5} & \textbf{R1@0.7} & \textbf{mAP@avg} & \textbf{R1@0.5} & \textbf{R1@0.7} & \textbf{mAP@avg}  \\
\midrule
30GB & 73.84 & 58.9 & 47.23 & 74.62 & 60.82  &  50.91 \\
35GB & 74.86 & 60.32 & 48.35 & 75.03 & 61.21 & 51.5 \\
40GB & 75.03 & 60.89 & 49.82 &77.74  &63.68 & 53.78  \\
45GB & 76.58 & 61.24 &51.22 & 78.84 & 64.19 & 54.25  \\
\bottomrule
\end{tabular}%
}
\caption{Performance comparison by memory usage. The SMORE model demonstrates superior performance over the frame-only baseline across all tested GPU memory usage budgets (MEM), achieving higher scores across all metrics.}
\label{tab:6}
\end{table}
\begin{figure}[h]
\centerline{\includegraphics[width=1\columnwidth]{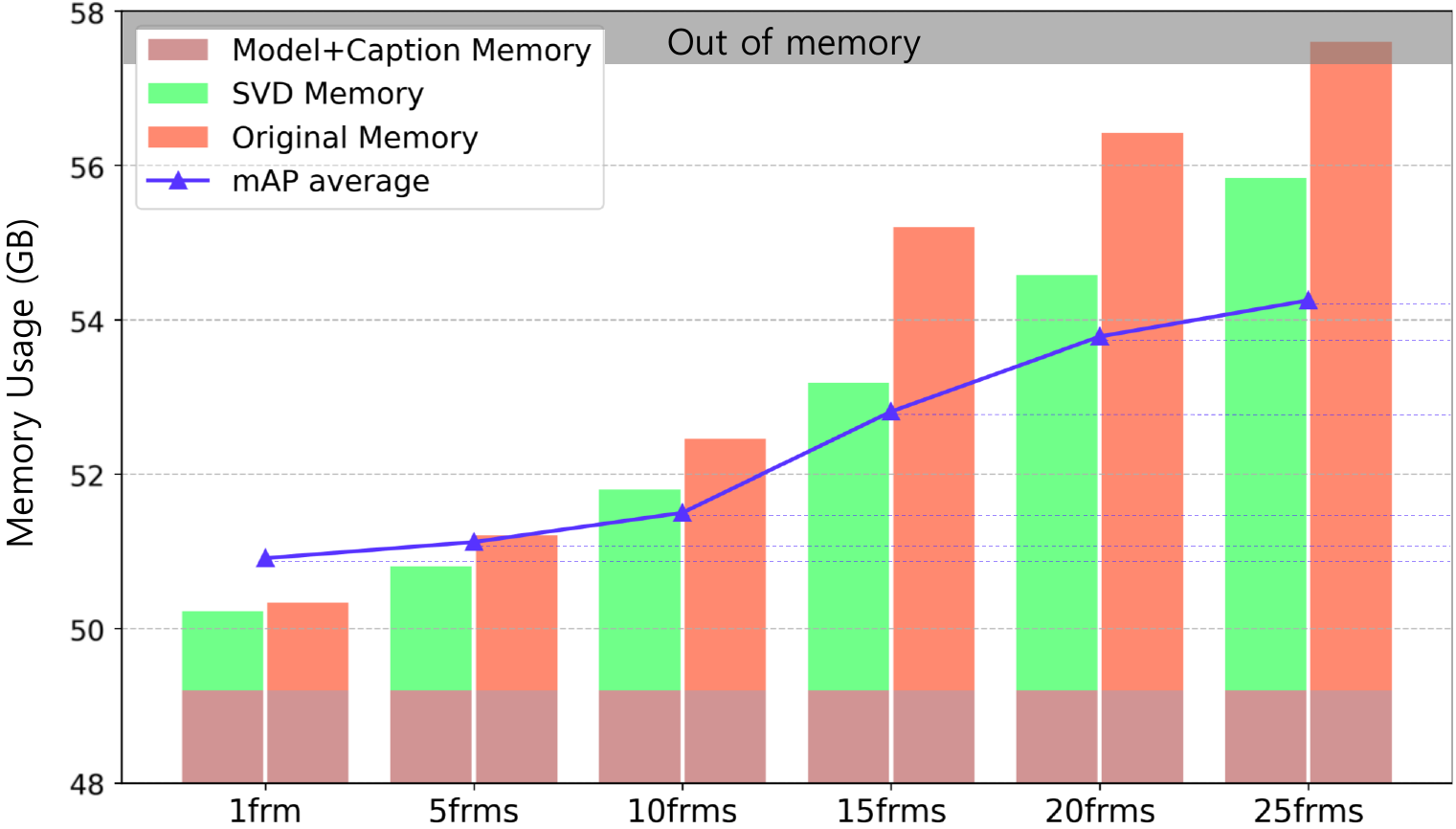}}
\caption{Variation in memory usage and performance of SMORE as a function of the number of sampled frames. Both memory usage and performance increase with more frames. However, our structured visual compression mitigates unnecessary computational overhead, contributing to improved memory efficiency.
}
\label{fig_6}
\end{figure}
In (d), the prediction was limited due to the inherent ambiguity of the query itself; for the query "A lady's video before the take-off of a plane", the ground truth should include all scenes before boarding the plane, but the model predicted starting from the interior of the plane. 
This outcome stems from an unclear query rather than a shortcoming of the model. Overall, SMORE demonstrates the effectiveness of its modules in qualitative evaluations.
Additionally, it exhibits strong performance on most datasets with clearly defined queries.

\subsection{Further Analysis}

\subsubsection{Memory Efficiency}
\label{Memory}
To demonstrate the memory efficiency of SMORE, we begin by comparing its memory usage against baseline models that operate directly on raw video frames.
As shown in Table~\ref{tab:6}, under the same memory constraints, SMORE outperforms the baselines across all metrics R1@0.5, R1@0.7, and mAP@avg.
Additionally, Figure~\ref{fig_6} presents a comparison of our model's performance and memory usage as a function of the number of sampled frames. The results demonstrate the memory efficiency of our method and further suggest that SMORE can achieve better performance in larger memory environments.
These results suggest that SMORE can scale to longer videos with richer multimodal information, achieving even stronger performance while maintaining memory efficiency.


\begin{table}[h]
\centering
\renewcommand{\arraystretch}{1}
\resizebox{0.95\columnwidth}{!}{%
\begin{tabular}{lcc}
\toprule
\textbf{Component} & \multicolumn{2}{c}{\textbf{Latency (s)}} \\
\cmidrule(lr){2-3}
& \textbf{SE Mode} & \textbf{LE Mode} \\
\midrule
QA-based Filtering (QAF)       & 2.79 & 2.79 \\
Query-guided Captioning (QGC)  & 1.47 & -    \\
Selective Re-captioning (SRC)  & -    & 0.51 \\
Importance Modulation (IM)     & 0.05 & 0.05 \\
\midrule
\textbf{Total Overhead}        & \textbf{$\sim$4.31} & \textbf{$\sim$3.35} \\
\bottomrule
\end{tabular}}
\caption{Latency breakdown for SMORE's operational modes. SE (Storage-Efficient) generates all captions on-demand, while LE (Latency-Efficient) uses pre-computation and selective re-captioning.}
\label{tab:latency}
\end{table}


\subsubsection{Latency and Practicality Analysis}
To demonstrate the efficiency of SMORE, we evaluate both its memory usage and query-time latency. As detailed in Table~\ref{tab:6} and Figure~\ref{fig_6}, SMORE consistently outperforms baseline models under identical memory constraints, confirming its high memory efficiency. Furthermore, our latency analysis in Table~\ref{tab:latency} reveals that SMORE is a flexible framework that can operate in two distinct modes to balance system priorities. The Storage-Efficient (SE) mode, which generates captions on-demand, shows a practical overhead of $\sim$4.31s. For applications where responsiveness is critical, the Latency-Efficient (LE) mode reduces this time to $\sim$3.35s by using pre-computed captions and selective re-captioning. Altogether, this proves that SMORE is not only a robust and memory-efficient solution but also a highly adaptable framework, suitable for diverse real-world deployments optimized for either storage or latency.



\section{Conclusion}

In this paper, we presented SMORE, a memory-efficient framework for Video Moment Retrieval that addresses the memory bottlenecks of MLLMs without compromising fine-grained temporal understanding. SMORE achieves this through two core strategies: (1) a query-guided semantic abstraction that refines textual representations to align with user intent, and (2) a structured visual compression that effectively reduces data redundancy.

These components collectively enable our model to achieve state-of-the-art performance across various benchmarks. The success of SMORE demonstrates that high retrieval accuracy and efficiency are not mutually exclusive, paving the way for deploying powerful video-language models on more accessible hardware.



\section{Limitations}
While our SMORE framework is based on an encoder-decoder architecture, a promising future direction is the exploration of decoder-only models for video understanding. This approach would allow for a wider range of potential architectures and more LLM-agnostic models, thereby extending the framework's applicability and generalizability.
Nevertheless, the current framework presents several challenges and opportunities for future improvement. First, the modules introduced to enhance both memory efficiency and accuracy create an inevitable trade-off in the form of a slight inference latency. Second, as observed in our qualitative analysis, a practical limitation exists where prediction accuracy degrades when faced with highly ambiguous videos or queries.
We believe these challenges can be effectively addressed through future work, focusing on pipeline optimization and strengthening the model's contextual reasoning capabilities.


\bibliography{custom}

\appendix
\section*{Appendix}
\label{sec:appendix}

\section{Implementation Details}
For Query-Aware Importance Modulation in Section 3.3, feature extraction for similarity comparisons was conducted using a CLIP-based model~\cite{CLIP}, and for the LLM, we utilized an encoder-decoder model, Flan-T5 XL~\cite{7_Flant5}. For query-guided caption generation (QA model), we employed InstructBLIP~\cite{8_InstructBLIP}, while BLIP2~\cite{blip2} was used for caption generation. The average length of captions generated by BLIP2 is 9.7 words. (To our knowledge, BLIP2 is the most suitable for SMORE; longer sentences would increase token count, complicating efficient segment sampling.)

For the SMORE module, considering the relatively small dataset size, we adopted a parameter-efficient fine-tuning method based on LoRA~\cite{LoRA}, training only 0.6266\% of the total model parameters. Additionally, to mitigate output instability inherent in large language models (LLMs), we applied a post-processing technique to correct formatting errors and enhance prediction accuracy. Other modules utilized in SMORE were used in a training-free manner.
For the Query-Aware Importance Modulation (Section 3.3), the weighting coefficients \(\alpha_1\) and \(\alpha_2\) were set to 0.7 and 0.3, respectively. The inter-frame similarity threshold \(\theta\) for Structured Visual Compression (Section 3.4) was set to 0.95.

Optimization was performed using AdamW~\cite{AdamW}. The learning rate was initially set at $1\times10^{-8}$, linearly increased to $3\times10^{-4}$ during the first 10\% of training, and then decayed following a cosine schedule. During training, frames were randomly sampled at uniform intervals. Tokens representing timestamps were rounded to the nearest integer to maintain memory efficiency as token counts increased. This optimization setup follows configurations from Chrono~\cite{13_chrono}.

Our dataset-specific experimental setups are as follows:
For QVHighlights and ActivityNet-Captions, we sampled 25 frames per video and generated one query-guided caption every 2 seconds. Training lasted up to 20 epochs with a batch size of 32, utilizing 8 A6000-48GB GPUs and a gradient accumulation step of 4. Total training time was about 93 GPU hours.
For Charades-STA, we sampled 30 frames per video and generated one query-guided caption every second. Training also lasted up to 20 epochs but used a batch size of 16, with 4 A6000-48GB GPUs and a gradient accumulation step of 4. Total training time was about 53 GPU hours.


\section{Hyperparameter Selection}

\subsection{Hyperparameter Ablation Study}
\label{sec:hyperparameter}


\begin{table}[h!]
\centering
\resizebox{\columnwidth}{!}{%
\begin{tabular}{ccccccc}
\toprule
\textbf{\(\alpha_1\)} & \textbf{\(\alpha_2\)} & \textbf{R1@0.5} & \textbf{R1@0.7} & \textbf{mAP@0.5} & \textbf{mAP@0.75} & \textbf{Avg.} \\
\midrule
0.9 & 0.1 & 77.92 & 63.17 & 68.89 & 54.88 & 53.65 \\
0.8 & 0.2 & 78.04 & 63.58 & 69.25 & 55.12 & 53.81 \\
\textbf{0.7} & \textbf{0.3} & \textbf{78.84} & \textbf{64.19} & \textbf{70.13} & \textbf{55.89} & \textbf{54.25} \\
0.6 & 0.4 & 78.52 & 63.83 & 69.90 & 55.63 & 54.01 \\
0.5 & 0.5 & 78.32 & 63.67 & 69.82 & 55.26 & 53.97 \\
\bottomrule
\end{tabular}}
\caption{Ablation study on the query-aware weighting coefficients \(\alpha_1\) and \(\alpha_2=1-\alpha_1\) on the QVHighlights validation set. The best performance is achieved at \(\alpha_1=0.7\), \(\alpha_2=0.3\).}
\label{tab:alpha_ablation}
\end{table}


\begin{table}[h!]
\centering
\resizebox{\columnwidth}{!}{%
\begin{tabular}{cccccc}
\toprule
\textbf{\(\theta\)} & \textbf{R1@0.5} & \textbf{R1@0.7} & \textbf{mAP@0.5} & \textbf{mAP@0.75} & \textbf{Avg.} \\
\midrule
0.85 & 77.20 & 62.90 & 69.80 & 55.20 & 53.78 \\
0.90 & 78.06 & 63.81 & 69.95 & 55.67 & 53.97 \\
\textbf{0.95} & \textbf{78.84} & \textbf{64.19} & \textbf{70.13} & \textbf{55.89} & \textbf{54.25} \\
0.99 & 78.06 & 63.53 & 68.12 & 55.24 & 53.89\\
\bottomrule
\end{tabular}%
}
\caption{Ablation study on the cosine similarity threshold \(\theta\) in the Structured Visual Compression module on the QVHighlights validation set. The optimal threshold is \(\theta=0.95\).}
\label{tab:theta_ablation}
\end{table}


In this section, we perform ablation studies to select the most effective hyperparameters for our model on the QVHighlights validation set. We first analyze the impact of the query-aware weighting coefficients \(\alpha_1\) and \(\alpha_2=1-\alpha_1\) by varying \(\alpha_1\) from 0.9 to 0.5 and measuring retrieval performance in terms of R1@0.5, R1@0.7, mAP@0.5, mAP@0.75, and the average score. We then investigate the effect of the cosine similarity threshold \(\theta\) in the Structured Visual Compression module by testing values between 0.85 and 0.99. Tables \ref{tab:alpha_ablation} and \ref{tab:theta_ablation} summarize these results, showing that \(\alpha_1=0.7,\,\alpha_2=0.3\) and \(\theta=0.95\) yield the best overall performance.






\end{document}